\documentclass[a4paper,10pt,twoside]{article}

\usepackage{clin}
\usepackage{harvard}
\usepackage[dutch, english]{babel}
\usepackage{graphicx}
\usepackage{enumerate}
\usepackage{caption}
\usepackage{subcaption}
\usepackage[title]{appendix}
\usepackage{url}		
\usepackage{hyperref}
\usepackage{mathtools}
\usepackage{array}

\newcolumntype{P}[1]{>{\centering\arraybackslash}p{#1}}

\hypersetup{
    hidelinks
}

\newcommand{\noopsort}[1]{}

\hypersetup{
    pdftitle={A Knowledge-Based Language Model},
    pdfauthor={David Ph. Shakouri, Crit Cremers, Niels O. Schiller},
    pdfkeywords={multiagent systems, modelling language acquisition, knowledge acquisition, statistical natural language processing, rule learning, language acquisition, graph theory for NLP, machine learning, logic-based artificial intelligence (AI), language models, function and content words}
}

\begin{document}

\title{A Knowledge-Based Language Model: Deducing Grammatical Knowledge in a Multi-Agent Language Acquisition Simulation}

\author{David Ph. Shakouri$^{*,**}$ \email{d.p.shakouri@hum.leidenuniv.nl}\\
{\normalsize \bf Crit Cremers}$^*$ \email{c.l.j.m.cremers@hum.leidenuniv.nl}\\
{\normalsize \bf Niels O. Schiller}$^{*,**,***}$ \email{Niels.Schiller@cityu.edu.hk}
\AND \addr{$^*$Leiden University Centre for Linguistics (LUCL), Leiden University, the Netherlands}
\AND \addr{$^{**}$Leiden Institute for Brain and Cognition (LIBC), Leiden University, the Netherlands} 
\AND \addr{$^{***}$City University of Hong Kong (CityU), Hong Kong} }

\maketitle\thispagestyle{empty}

\jmlrheading{14}{2025}{167-189}{12/2024}{07/2025}{David Ph. Shakouri, Crit Cremers, and Niels O. Schiller}

\begin{abstract}
This paper presents an initial study performed by the MODOMA system. The MODOMA is a computational multi-agent laboratory environment for unsupervised language acquisition experiments such that acquisition is based on the interaction between two language models, an adult and a child agent. Although this framework employs statistical as well as rule-based procedures, the result of language acquisition is a knowledge-based language model, which can be used to generate and parse new utterances of the target language. This system is fully parametrized and researchers can control all aspects of the experiments while the results of language acquisition, that is, the acquired grammatical knowledge, are explicitly represented and can be consulted. Thus, this system introduces novel possibilities for conducting computational language acquisition experiments. The experiments presented by this paper demonstrate that functional and content categories can be acquired and represented by the daughter agent based on training and test data containing different amounts of exemplars generated by the adult agent. Interestingly, similar patterns, which are well-established for human-generated data, are also found for these machine-generated data. As the procedures resulted in the successful acquisition of discrete grammatical categories by the child agent, these experiments substantiate the validity of the MODOMA approach to modelling language acquisition.
\end{abstract}

\section{Introduction}
\label{sec:Introduction}
This paper presents a study demonstrating the acquisition of grammatical knowledge by the MODOMA. The term MODOMA is an acronym for \textit{moeder-dochter-machine} (Dutch for ‘mother-daughter-machine’). This framework is aimed at providing a language acquisition laboratory, that is, a simulation environment for language acquisition experiments. On the one hand, all aspects of the system are parametrized so that the settings are user-controlled while on the other hand, all results and executed language acquisition procedures are logged and can be retrieved. The MODOMA integrates several characteristics that enable unique possibilities for language acquisition experiments. The MODOMA implements a multi-agent design modelling parent-child interaction such that both the parent and the child are language models in a single system. Both agents employ explicit representations of their grammatical knowledge, making the acquired knowledge and language processing retrievable. This explicit representation distinguishes the MODOMA system from other language learning systems, such as large language models (LLMs), which do not rely on such explicit knowledge structures. This combination of properties provides new opportunities for conducting computational experiments simulating first language acquisition. In a typical MODOMA experiment the input data to the language acquisition algorithm are interactively generated by the mother agent while the daughter agent constantly updates grammatical representations and takes part in the interaction with the mother agent based on the currently acquired grammar. This design represents a novel approach as most often computational models of language acquisition are based on inputting corpora to the language acquisition procedures \citeaffixed{Alishahi:2008,Alishahi:2012,Conner:2009,Matusevych:2013}{e.g.}.

In this paper, rather than limiting the term "language model" to the typical usage in current NLP, where it usually refers to neural models trained on a next-word prediction objective (e.g., transformer-based models like GPT, cf. \citename{Brown:2020} 2020, \citename{Radford:2018} 2018, and BERT, cf. \citename{Devlin:2019} 2019), we use it in a broader sense, encompassing rule-based approaches, statistical methods, and artificial neural networks as all of these approaches are employed to address common tasks in NLP. This distinction is important because the mother agent in our multi-agent system relies on explicit linguistic rules, while language acquiring agent incorporates statistical methods to infer a rule-based model for a linguistic phenomenon.

The study presented by this paper investigates whether the MODOMA daughter agent is able to acquire and represent functional and content categories by integrating an application of the Zipfian distribution in the MODOMA language acquisition system. Specifically, we examine whether a multi-agent language acquisition system can derive functional and content categories from machine-generated utterances and represent them through a rule-based grammar. To address this question, we assess whether these utterances exhibit statistical patterns similar to those in human language and determine the amount of input data necessary to detect such patterns. We then explore whether these statistical tendencies can be converted into grammatical rules. Finally, we validate the results by assessing the alignment of the inferred categories with the original grammar and evaluating the system’s reliability in acquiring structured linguistic knowledge in an unsupervised manner.

The discussed language acquisition experiments are aimed at adding explicit grammatical knowledge to a lexicon. Subsequently, the daughter agent can use this acquired  knowledge to take part in the conversation with the mother. To this end, the system employs statistical machine learning techniques, resulting in an explicitly specified grammatical representation. This contributes to existing systems based on statistical methods and neural networks by offering a more interpretable, rule-based approach to language modeling. By employing the principle of Zipf's law \cite{Mandelbrot:1954,Zipf:1949}, this study provides an essential robustness evaluation of the feasibility of implementing a multi-agent simulation of first language acquisition. Crucially, the purpose of this study is not to perform a binary classification of function and content words per se, but rather to employ this classification as a tool for assessing the validity of this novel approach to modelling first language acquisition. The advantage of using Zipf's law is that it offers a solid and reliable basis for evaluating the system’s performance  and components. This is a well-established practice. For example, \citeasnoun{Chaabouni:2019b} and \citeasnoun{Rita:2020b} employ Zipf’s law to develop a multi-agent framework to model language evolution, see also the recent studies by \citeasnoun{Debowski:2021}, \citeasnoun{Takahashi:2019}, and \citeasnoun{Takamichi:2024}.

Importantly, the experiments in this study validate all components necessary to enable a multi-agent model of first language acquisition. Specifically, the acquisition of function and content words employs a hybrid language acquisition strategy, which is based on statistical tendencies that facilitate the acquisition of discrete grammatical rules. Therefore, these results support the modeling of more complex grammatical phenomena and demonstrate the feasibility of the MODOMA research agenda. A study building on these results is the work by \citeasnoun{Shakouri:2025}, which demonstrates how grammatical categories such as noun, verb, and preposition can be acquired using unsupervised techniques. In this paper two experiments illustrating this approach are presented: the first simulation discussed in Sections \ref{sec:Results} and \ref{sec:Analyses} has been applied to a training data set generated by the mother agent. These experiments were used to determine the parameter settings for the acquisition of this linguistic phenomenon. The second experiment taking a novel test data set generated by the adult language model as its input was used to validate this configuration, see Section \ref{sec:Testing}. The results suggest that the MODOMA is able to acquire these grammatical categories and represent them by the grammar formalism of the daughter agent.

\section{Description of the MODOMA}
\label{sec:Description_MODOMA}

\newlength{\lenA}
\settowidth{\lenA}{\textbar SLF: }

\begin{figure*}
\centering
\begin{tabbing}
ID:A+B\\
\textbar HEAD: \=\textbar CONCEPT:the\\
\textbar \> \textbar PHON:de\\
\textbar \> \textbar QSEM:the\\
\textbar \> \textbar SLF:the\\
\textbar \> \textbar SYNSEM:CAT:det\\
\textbar PHON:C\\
\textbar PHONDATA:lijnop(de,A+B,[arg(right(1),0,D)],C)\\
\textbar SLF: \parbox{\dimexpr\textwidth-\lenA\relax}{\{\{[E\$F\&(B+G)\#H],[\hspace*{0.5em}],[\hspace*{0.5em}]\},I@some\textasciicircum F\textasciicircum  and(and(quant(F,the),\break and(H,entails1(F,decr))),and(I,entails(F,incr)))\}}\\
\textbar SYNSEM: \=\textbar CAT:np\\
\textbar \> \textbar EMBEDCONCEPT:J\\
\textbar \> \textbar EXAGGR:[number:plur,person:3]\\
\textbar \> \textbar EXTTH:K~[A+B,L]\\
\textbar \> \textbar FREEARGS:M\\
\textbar \> \textbar FUNCL:decr\\
\textbar \> \textbar FUNCR:incr\\
\textbar \> \textbar GENDER:N\\
\textbar \> \textbar NOMTYPE:O\\
\textbar \> \textbar NUMBER:plur\\
\textbar \> \textbar PERSON:3\\
\textbar \> \textbar QMODE:def\\
\textbar \> \textbar REFMODE:P\\
\textbar \> \textbar SEX:Q\\
\textbar \> \textbar SUBCAT:noun\\
\textbar \> \textbar VOICE:R\\
\textbar TYPE:np\textbackslash0$\sim$[\hspace*{0.5em}]/0$\sim$[n\textasciicircum0\#B+G]\\
\textbar ARG: \=\textbar ID:B+G\\
\textbar \> \textbar HEAD:CONCEPT:J\\
\textbar \> \textbar PHON:D\\
\textbar \> \textbar SLF:E\\
\textbar \> \textbar SYNSEM: \=\textbar CAT:n\\
\textbar \> \textbar \> \textbar EXTTH:K~[B+G,L]\\
\textbar \> \textbar \> \textbar FREEARGS:M\\
\textbar \> \textbar \> \textbar GENDER:N\\
\textbar \> \textbar \> \textbar NOMTYPE:O\\
\textbar \> \textbar \> \textbar NUMBER:plur\\
\textbar \> \textbar \> \textbar REFMODE:P\\
\textbar \> \textbar \> \textbar SEX:Q\\
\textbar \> \textbar \> \textbar VOICE:R\\
\end{tabbing}
\caption{Example of the representation of the lexical item for \textit{de} (`the.\textsc{masc/fem}') by Delilah}
\label{fig:Lexical_item_Delilah}
\end{figure*}

The MODOMA consists of two main components: (1) the mother agent and (2) the daughter agent. Delilah \cite{Cremers:1995}, the Leiden generator and parser of Dutch, is employed as the mother. Crucially, Delilah’s grammatical knowledge is not based on a corpus. The parser as well as the generator of the mother language model execute the same grammar consisting of predefined graph structures, which are comprised of attribute-value matrices. These graphs correspond to words and grammatical constructions and are explicitly specified for many grammatical properties such as the phonological form, concepts, logical meaning representation \citeaffixed{Cremers:2008}{cf.}, grammatical number, grammatical person, and the syntactic category (e.g. determiner, noun, or verb). 
An example of a Delilah template is provided in Figure \ref{fig:Lexical_item_Delilah} illustrating Delilah's lexical and grammatical knowledge of the word \textit{de} (`the.\textsc{masc/fem}’)\footnote{Appendix \ref{appendix:A} provides a glossary of grammatical abbreviations.}.  These graphs have been specified based on linguistic analyses of the Dutch lexicon and grammar, making Delilah's grammar closely tied to linguistic insights into Dutch. Thus, the linguistic knowledge and processing can be consulted. In this respect Delilah offers a complementary perspective, particularly when contrasted with the data-driven approach of large language models.  

While Delilah’s lexicon contains a large but finite list of words and constructions, these templates can be combined to form more complex utterances such as noun phrases and sentences. For example, the template for the article \textit{de} (`the.\textsc{masc/fem}’’) can be used by Delilah to form noun phrases (e.g., \textit{de vrouw}, `the woman’). These templates are combined through the process of unification. This is a mathematical procedure, which verifies whether the templates contain conflicting features. If no conflicting values are found, it creates a template specified for all information contained in the original templates, see \citeasnoun{Schieber:2003} for a more detailed account of unification applied to grammatical representations. As shown in Figure \ref{fig:Lexical_item_Delilah}, the complement position associated with the article \textit{de} (`the.\textsc{masc/fem}’) is constrained to only contain noun phrases, as indicated by the specification ARG\textbar SYNSEM\textbar CAT:n. Given the highly detailed nature of Delilah’s representations, it lies beyond the scope of this paper to provide a full account of all features involved. However, \citeasnoun[259-265]{Cremers:2014} provide an elaborate account of unification in the context of Delilah. In the application of unification to graph structures to produce and parse utterances, Delilah resembles head-driven phrase structure grammar (HPSG, \citename{Pollard:1986} 1986, \nocite{Pollard:1994} 1994). Here the terms ‘grammar’ and ‘lexicon’ are used synonymously as there is no principled formal distinction between more concrete words and more abstract grammatical rules. Crucially, this language model executes a combinatory list grammar, which is related to combinatory categorial grammar (CCG, \citename{Cremers:2002} 2002, pp.~378-386, \citename{Cremers:2014} 2014, pp.~115-137, cf. \citename{Baldridge:2003} 2003, \citename{Moortgat:1997} 1997, \citename{Steedman:1996} 1996). The characteristics of this language model are extensively discussed by \citeasnoun{Cremers:2014} and \citeasnoun[pp.~25-75]{Reckman:2009}.\footnote{A demo version can be consulted at \url{https://delilah.universiteitleiden.nl/indexen.html} \cite{Cremers:2024}, last accessed Oct. 1, 2024.} For the purposes of the MODOMA, Delilah, the mother agent, has been taken as is.

Conversely, the daughter language model has been constructed specifically as part of the MODOMA. Similar to Delilah, the daughter agent is a parser and generator but unlike the adult language model the grammatical representations are underspecified and become more specific as a result of the language acquisition procedures. A non-trivial resemblance between the adult and child systems is that both employ graph structures for representing grammatical knowledge and their parsers and generators execute their respective grammars by unifying lexical structures to output well-formed utterances and provide grammaticality judgements and parses with respect to input utterances based on the currently specified grammar. A crucial characteristic of the parser and generator of the child agent is that they can execute an incomplete grammar containing underspecified structures as similar to humans acquiring their first languages the daughter language model is required to take part in the conversation with the mother before acquisition has completed.

\begin{figure*}[t]
	\centering
	\[
		\begin{bmatrix*}[l]
			\text{memory stack position:} & \text{6878}\\
			\text{lexical entry number:} & \text{6879}\\
			\text{session ID:} & \text{1731286472057}\\
			\text{confidence lexical entry:} & \text{60}\\
			\text{head directionality:} & \text{null}\\
			\text{confidence head directionality:} & \text{0}\\
			\text{terminal:} & \text{T}\\
			\text{phonform:} & \textit{auto}\\
			\text{semform:} & \text{EKC}\\
			\text{semform index:} & \text{null}\\
			\textsc{grammatical properties:} &
			\begin{bmatrix*}[l]
				\text{property type:} & \textsc{A}\\
				\text{property value:} & \text{b}\\
				\text{confidence property:} & \text{60}\\
			\end{bmatrix*}\\
				\textsc{head:} & \begin{bmatrix*}[l] \textsc{auto.6879} \end{bmatrix*}\\
				\textsc{argument:} & \begin{bmatrix*}[l] \text{null} \end{bmatrix*}\\
		\end{bmatrix*}
	\]
	\caption{Example of the representation of an acquired lexical item for \textit{auto} (‘car’) by the daughter agent}
	\label{fig:Lexical_item_daughter}
\end{figure*}		

The daughter agent comprises three modules: (1) a parser and generator, the automaton, which execute (2) a grammar, which is specified as a result of acquisition by (3) a language acquisition device \citeaffixed{Shakouri:2018}{cf.}. Crucially, the parser and generator are not modified by the language acquisition procedures. The result of language acquisition is explicit grammatical knowledge represented by newly acquired lexical entries and/or grammatical properties of lexical entries limiting their combinatory properties. A minimal example of a lexical entry is given by Figure \ref{fig:Lexical_item_daughter}. These data structures provide comprehensive representations of words and constructions. They have been specified for several grammatical properties, for example the phonological form (phonform), a representation of semantic content (semform), and an expandable list of other acquired grammatical properties. Grammatical properties are formalized by feature-value pairs such as [\textsc{Content or function word}: content word] and [\textsc{Content or function word}: function word] the acquisition of which is the aim of the study presented by this paper. As the learning procedures are unsupervised, the daughter language model has no information on the labels employed by the mother language model or grammar descriptions. Therefore, the acquired grammatical properties and the semform are indicated by unique alpha-numeric labels such as the property type ‘A’ and property value ‘b’ in Figure \ref{fig:Lexical_item_daughter}, see also Figure \ref{fig:Repres_Cat}. These representations encode other grammatical properties not investigated by this paper as well. For example, the terminal feature differentiates between terminal words and more abstract constructions while the semform index enables representing anaphora such as reflexives. In addition, technical features are included such as the session id and the lexical entry number, which uniquely identify each rule. As a result of simulating language acquisition, the grammar can consist of increasingly complex structures as the \textsc{head} and \textsc{argument} can contain data structures identical to lexical items.

Since all linguistic knowledge is explicitly represented and can be consulted by researchers with respect to both agents, this type of approach provides an insightful addition to systems that learn language but do not employ explicit representations such as large language models. The MODOMA implements unsupervised learning as the input data to the language acquisition device are unannotated and this module has no access to the mother agent’s internal processing and parses. Moreover, previously acquired grammatical knowledge by the daughter agent can be employed as input to the language acquisition procedures. This additional acquisition technique, which was introduced for the purposes of the MODOMA, has been named internal annotation. This is a form of self-supervised learning \citeaffixed{Balestriero:2023,Brown:2020,Devlin:2019,Gui:2024,Lan:2020,Orhan:2020,Yarowsky:1995}{cf.}.

\section{Related work}
\label{sec:Related_work}
As far as we know, the combination of characteristics and research possibilities entailed by the multi-agent MODOMA framework is a novel design. In particular, the resulting system can be used to cast language acquisition experiments based on natural language involving an adult language model and an acquiring language model. These agents take part in interactions, such that all aspects of the system, for example the parameter settings, input, executed language acquisition procedures, interaction, generated utterances, and acquired grammar, are user-controlled, reproducible, measurable, and verifiable. Ter Hoeve et al. \citeyear{Hoeve:2022} present a road map of a teacher-student-loop to model language acquisition and implement two experiments corresponding to the first steps. The first experiment is aimed at modelling separate domains by ‘two strictly separated vocabularies’ of an artificial language while the second experiment addresses distinctive structures modeled by repetitions of tokens. As part of these experiments, the teacher selects a fixed number of training data from a set of pre-constructed sentences to provide them to multiple students, which consist of language models. After training, the student language models are examined with respect to a test set consisting of different sentences of the artificial language. In addition, a teacher language model is trained on another subset of sentences and subsequently assessed. The current implementation of this framework differs from the MODOMA most notably with respect to the training and test data: In the case of the work by \citeasnoun{Hoeve:2022} the input contains pre-constructed utterances of an artificial language while for the MODOMA the data are representative of a natural language and produced online by Delilah as part of the interaction based on an explicitly defined grammar model of Dutch.

From an evolutionary perspective, research has been conducted on emergent communication utilizing multi-agent systems. An important example is the Talking heads framework. This project has been developed by \citename{Steels:2000} \citeyear{Steels:2000,Steels:2015} and provides a model of language evolution based on interactions including references to non-linguistic contexts between agents. These agents develop an artificial language as a result of language games \cite{Steels:2001} such as the naming game aimed at defining names for objects in the surroundings (cf. \citename{SteelsKaplan:2001} 2001 for studies on word naming, \citename{Steels:2012} 2012). Currently, their work also addresses the emergence and acquisition of syntactic phenomena \citeaffixed{Steels:2017,Steels:2018,Trijp:2016}{cf.}. Relevant differences with the MODOMA project are that both agents engage in developing and learning language. Moreover, while the MODOMA daughter addresses the acquisition of (a fragment of a) natural language, the talking heads acquire an artificial language.

\citeasnoun{Chaabouni:2020b} study the emergence and beneficial properties of compositionality and generalization by two deep neural agents: A Receiver agent constructs a message \textit{m} based on a received input \textit{i} while a Sender agent is inputted with \textit{m} and returns an output \textit{\^{i}}. Then, the game is successful if the input matches the output, that is, \textit{i = \^{i}}. \citeasnoun{Chaabouni:2022} provide an architecture involving a Speaker, which receives an image and outputs a message, and a Listener, which aims at selecting the same image from a set of images based on the message, to study the effects of scaling up the ‘data set, task complexity, and population size’. Although these studies involve two agents, unlike the MODOMA the language employed by these agents involves ‘emergent codes’ rather than (a fragment of) a natural language. Interestingly, \citeasnoun{Chaabouni:2019b} investigate whether two communicating agents can create an artificial language following Zipf's law, namely that more frequent words are likely to be shorter. To the contrary, they demonstrate that an anti-efficient code arises such that the most frequent words are correlated with the longest messages. Conversely, \citeasnoun{Rita:2020b} indicate an artificial language congruent with Zipf’s law is learned for a LazImpa system involving a ‘Lazy Speaker’ and an ‘Impatient Listener’. \citeasnoun{Griffith:2007} model language evolution by (a population of) Bayesian agents executing iterative learning with respect to languages consisting of combinations of representations of meanings and utterances.

Alishahi carried out research simulating the acquisition of language by applying algorithms to corpora that have been human or artificially generated, see \citename{Alishahi:2009} \citeyear{Alishahi:2009,Alishahi:2012} and \citeasnoun{Alishahi:2008} for studies modelling the acquisition of lexical categories, word meaning, and early argument structure. \citeasnoun{Matusevych:2013} enhance artificially generated input corresponding to child-adult interactions to use them similarly to data such as the CHILDES database \cite{MacWhinney:2014} but differently from the MODOMA framework as the algorithm that executes language acquisition, does not take part in the conversation. \citeasnoun{Beekhuizen:2014} provide a different perspective by modelling the acquisition of grammatical constructions employing a single agent taking as its input generated representations of utterances and meanings. Furthermore, \citename{Conner:2008} \citeyear{Conner:2008,Conner:2009}’s Baby SRL project models the acquisition of semantic role labelling by inputting annotated data from the CHILDES corpus or constructed sentences into their algorithm.

\section{Modelling the acquisition of functional and content categories}
\label{sec:Modelling_acquisition}
The experiments presented in this paper assess whether a computational laboratory simulation of language learning could acquire function  and content categories in an unsupervised fashion as a result of an interaction with Delilah, the language model providing samples of the adult language. Subsequently, the daughter language model can employ this newly acquired grammatical knowledge to take part in an interaction with the mother. On the one hand, the daughter agent can use the acquired grammar specified for these classifications to generate new sentences. These sentences can be presented to the mother as part of the conversation and the mother agent can return feedback (if requested by the parameter settings of the MODOMA) so that the daughter can assess the quality of the classifications. On the other hand, during the interaction the daughter uses this grammatical knowledge to parse novel utterances produced by the mother, for instance by classifying words presented by the mother with respect to this distinction.

Interestingly, it is a well-known finding in corpus linguistics that words corresponding to functional categories tend to occur with a high frequency whereas content words tend to have a low frequency of occurrence: many content words are used as hapax legomena, that is, only once, or a few times in a text, see \citeasnoun{Powers:1998} for an analysis of closed and open class words. This finding is congruent with Zipf’s law, that is, the observation that the ranking of a word in a list of mostly used words is (approximately) proportional to the frequency of use for each word with respect to a particular factor \citeaffixed{Mandelbrot:1954,Zipf:1949}{cf.}. Moreover, a correlation between a distribution observed in languages and grammatical categories distinguished by linguists is suggested for language produced by humans. This paper assesses whether a similar correlation can be found for language produced by a machine. 

An initial experiment implemented in the MODOMA laboratory simulation environment has been based on this correlation. This experiment provides a method to evaluate the MODOMA approach to language acquisition experiments employing the well-established principle of Zipf's law. Thus, this study contributes to a research development such that the MODOMA is gradually evolved by implementing increasingly complex experiments. As a result, experience with the MODOMA as a laboratory model for language acquisition experiments can be increased and the system can be tested and improved before implementing additional language acquisition techniques. Although \citeasnoun{Chaabouni:2019b} and \citeasnoun{Rita:2020b} focus on language emergence, there are some similarities with respect to the application of Zipf's law to increase understanding of language models in comparison with human-generated languages. Crucially, this procedure involves most of the components and procedures required by more intricate acquisition experiments. Accordingly, the component parts of the MODOMA system and their interaction, can be validated and improved to enable the implementation of further computational language acquisition experiments. If the acquisition succeeds, a distinction between functional and content categories is learned by the daughter agent based on samples of the target language presented by the mother agent during an interaction with the daughter. In this respect, it is significant that as soon as the daughter agent has acquired this grammatical distinction, the new grammatical knowledge is immediately used to update the daughter grammar and subsequently employed to generate novel utterances during the conversations with the mother agent and parse novel mother sentences. Consequently, this study provides an important indication of whether the MODOMA laboratory approach to language acquisition can result in the acquisition of non-trivial grammatical knowledge that can be used productively during parsing and generation.

\section{Methodology}
\label{sec:Methodology}

\begin{table}[h]
\centering
\begin{tabular}{p{4,6cm}p{10,2cm}}
\hline
\textbf{Parameter} & \textbf{Description}\\
\hline
Acquisition of functional categories & determines whether the system attempts to execute the acquisition of functional and content categories.\\\hline
type of phrase generated by the mother agent & specifies whether Delilah should generate noun phrases (NPs) or full sentences when requested to produce utterances.\\\hline
Minimum amount of data processed by the daughter agent & defines the minimum amount of data the system should process before starting the acquisition of functional and content categories.\\\hline
Threshold for functional category acquisition & defines the threshold based on the type/token ratio to differentiate between functional and content categories in the utterances generated by the mother agent.\\
\hline
\end{tabular}
\caption{Parameters for the functional and content category acquisition experiments}
\label{tab:parameters}
\end{table}

As the MODOMA implements a laboratory environment for language acquisition experiments, it is a fully parametrized system. This means that users can control the properties of each experiment by specifying predefined settings. Crucially, all aspects of language acquisition, that is, the adult, the daughter and the executed language procedures, are included in the system. Accordingly, the parameters can pertain to all these components. Table \ref{tab:parameters} provides an overview of the relevant parameters for the current experiments. The MODOMA is designed to execute several acquisition procedures (see also \citename{Shakouri:2025} 2025, for an example of another acquisition procedure implemented using the MODOMA). The first parameter in Table \ref{tab:parameters}, \textit{Acquisition of functional categories}, specifies whether the acquisition of functional categories is enabled. In all experiments discussed in this paper, this is the case.

The remaining parameters define key experimental properties. The \textit{type of phrase generated by the mother agent} parameter controls the structure of Delilah’s utterances, determining whether sentences or noun phrases are generated. The training and test experiments presented in this paper systematically vary this parameter. Similarly, the \textit{minimum amount of data processed by the daughter agent} before acquiring functional and content categories is varied across all experiments. These two parameters define the experimental setup. Specifically, in the training experiments, acquisition of functional and content categories is triggered after processing 1,000 noun phrases (NPs) and 1,000 sentences, followed by 10,000 noun phrases (NPs) and 10,000 sentences, respectively. This allows us to assess how many input exemplars generated by Delilah are required to detect this statistical tendency. Additionally, the \textit{threshold for functional category acquisition} parameter specifies the type-token ratio used to convert distributional differences between function and content words into discrete linguistic knowledge. The training sessions discussed in this paper are used to determine appropriate values for these parameters, while the test experiments verify the selected settings using new datasets generated by Delilah.

\begin{figure*}
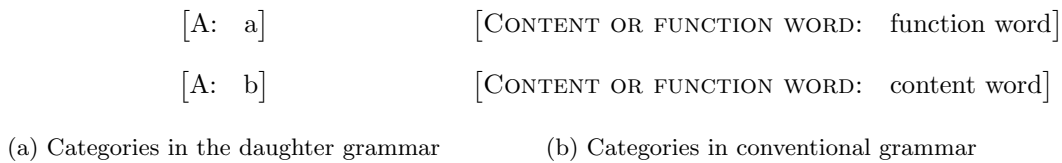

\centering
\begin{subfigure}{.38\linewidth}
	\centering
	\[
		\begin{bmatrix*}[l]
			 \text{A:} & \text{a}
		\end{bmatrix*}
	\] \[
		\begin{bmatrix*}[l]
			\text{A:} & \text{b}
		\end{bmatrix*}
	\]
	\caption{Categories in the daughter grammar}
	\label{fig:Cat_MODOMA}
\end{subfigure}\hspace{0,25cm}
\begin{subfigure}{.5\linewidth}
	\centering
	\[
		\begin{bmatrix*}[l]
		\textsc{Content or function word:} & \text{function word}
	\end{bmatrix*}
	\] \[
		\begin{bmatrix*}[l]
		\textsc{Content or function word:} & \text{content word}
		\end{bmatrix*}
	\]
	\caption{Categories in conventional grammar}
	\label{fig:Cat_trad_grammar}
\end{subfigure}
\caption{Representation of categories in the MODOMA versus conventional grammatical terminology}
\label{fig:Repres_Cat}
\end{figure*}

Accordingly, Delilah was requested to generate eight sets of respectively 1,000 and 10,000 NPs and sentences. For each experiment, two of these sets were used to configure the acquisition procedures while two other data sets were employed to assess whether based on the same parameter settings similar results could be obtained. Initially, an experiment has been executed using the training sets of 1,000 NPs and sentences to determine whether diverging distributions between function and content words as established by the literature for human-generated corpora \citeaffixed{Powers:1998}{cf.} could be obtained for utterances machine-generated by Delilah and whether this distinction is sufficiently attested for the purposes of acquiring grammatical features. Subsequently, a follow-up experiment was performed with respect to the training and test data sets containing 10,000 NP and sentence exemplars produced by Delilah. If this distinction is detected by the language acquisition procedures in such a way that words can be reliably classified, the results are represented using graphs consisting of feature-value pairs specified by alpha-numeric labels, which should correspond to terminology more conventionally employed by linguists, compare Figure \ref{fig:Cat_MODOMA} with example labels applied by the MODOMA daughter agent and Figure \ref{fig:Cat_trad_grammar} indicating more conventional labels.

\section{Results}
\label{sec:Results}
\begin{table}[b]
\centering
\begin{tabular}{lP{4cm}P{4cm}}
\hline
& \parbox[c][0.9cm][c]{4cm}{\centering \hspace{0.7cm} \textbf{NP}\newline \hspace{0.2cm} \textbf{training experiment}} & \parbox[c][0.9cm][c]{4cm}{\centering \hspace{0.7cm} \textbf{Sentence}\newline \hspace{0.2cm} \textbf{training experiment}} \\
\hline
\textbf{Number of word types} & 1,319 & 1,728\\
\textbf{Number of word tokens} & 3,027 & 5,874\\
\textbf{Number of hapax legomena} & 776 & 859\\
\textbf{Number of word types used twice} & 309 & 366\\
\hline
\end{tabular}
\caption{Lexical statistics of the 1,000 generated NP and sentences training data}
\label{tab:lex_stat_train_1000}
\end{table}

Table \ref{tab:lex_stat_train_1000} presents lexical statistics of the training data containing 1,000 generated noun phrases and sentences by Delilah, including the total number of unique word types, the total number of word tokens, the number of hapax legomena (words that occur only once), and the number of word types that appear exactly twice. As an intermediate step during the acquisition procedures, for each experiment the word types have been sorted by frequency, see Table \ref{tab:1000NP_training} and Table \ref{tab:1000S_training}.\footnote{In these tables, $f$ stands for frequency, that is, the number of tokens detected by the language acquisition procedures in the input data generated by the adult agent.} Interestingly, although this sample of 1,000 Delilah sentences contains more words than the collection of 1,000 NPs, the first unequivocal content word occurs at a higher rank in the list for 1,000 sentences, that is, \textit{vergeten} (‘forgotten’) at rank 9. In accordance with the research hypothesis, the initial sets of 1,000 exemplars indicate a distinction between functional and content categories considering the most frequent words most often correspond to functional categories. However, this distinction is not clear enough for a system designed to acquire grammatical categories. Therefore, it is concluded that although data sets containing 1,000 exemplars generated by Delilah exhibit a distinction between function and content words similar to the one found for human language output, this difference is not substantiated sufficiently to enable the acquisition of grammatical features.

\begin{table*}[h]
\begin{minipage}[l]{.4\linewidth}
\centering
\begin{tabular}[t]{@{}lll@{}}
\hline
\# & Word type & $f$ \\ \hline
1. & \textit{te} (‘at’, ‘to’, \textsc{comp}) & 92 \\
2. & \textit{dat} (‘that’) & 62 \\
3. & \textit{niet} (‘not’) & 60 \\
4. & \textit{om} (‘to’) & 49 \\
5. & \textit{geen} (‘no’) & 33 \\
6. & \textit{er} (‘there’) & 32 \\
7. & \textit{de} (‘the’, \textsc{masc/fem/pl}) & 25 \\
8. & \textit{elke} (‘every’, \textsc{masc/fem}) & 25 \\
9. & \textit{een} (‘a’, ‘one’) & 24 \\
10. & \textit{het} (‘the’, \textsc{neut:sg}) & 24 \\
11. & \textit{op} (‘on’) & 23 \\
12. & \textit{ieder} (‘every’, \textsc{neut}) & 21 \\
13. & \textit{ik} (‘I’) & 21 \\
14. & \textit{die} (‘that’, \textsc{masc/fem}) & 20 \\
15. & \textit{elk} (‘every’, \textsc{neut}) & 20 \\
16. & \textit{iedere} (‘every’, \textsc{masc/fem}) & 20 \\
17. & \textit{veel} (‘a lot of’) & 20 \\
18. & \textit{alle} (‘all’) & 19 \\
19. & \textit{daar} (‘there’) & 18 \\
20. & \textit{deze} (‘this’, \textsc{masc/fem}) & 17 \\
21. & \textit{hier} (‘here’) & 17 \\
22. & \textit{of} (‘or’, ‘whether’) & 17 \\
23. & \textit{voor} (‘for’) & 17 \\
24. & \textit{aan} (‘to’, \textsc{prep}) & 16 \\
25. & \textit{massa's} (‘masses’) & 15 \\
26. & \textit{van} (‘of’) & 14 \\
27. & \textit{vier} (‘four’) & 14 \\
28. & \textit{weinig} (‘a few’) & 14 \\
29. & \textit{dit} (‘this’, \textsc{neut}) & 12 \\
30. & \textit{menig} (‘many’, \textsc{sg}) & 12 \\
31. & \textit{negenennegentig} (‘ninetynine’) & 12 \\
32. & \textit{blijkt} (‘turns out’) & 11 \\
33. & \textit{naar} (‘to’, \textsc{prep}, ‘unpleasant’) & 11 \\
34. & \textit{twaalf} (‘twelve’) & 11 \\
35. & \textit{vijf} (‘five’) & 11 \\ \hline
\end{tabular}
\caption{35 most frequent words for 1,000 generated NPs}
\label{tab:1000NP_training}
\end{minipage}\hfill\begin{minipage}[l]{.4\linewidth}\centering
\begin{tabular}[t]{@{}lll@{}}
\hline
\# & Word type & $f$ \\ \hline
1. & \textit{dat} (‘that’) & 197 \\
2. & \textit{te} (‘at’, ‘to’, \textsc{comp}) & 193 \\
3. & \textit{er} (‘there’) & 119 \\
4. & \textit{niet} (‘not’) & 91\\
5. & \textit{ik} (‘I’) & 75 \\
6. & \textit{hebben} (‘have’) & 60 \\
7. & \textit{zijn} (‘are’, ‘be’, ‘his’) & 50 \\
8. & \textit{om} (‘to’) & 48 \\
9. & \textit{vergeten} (‘forgotten’) & 41 \\
10. & \textit{alle} (‘all’) & 38 \\
11. & \textit{deze} (‘this’, \textsc{masc/fem}) & 38 \\
12. & \textit{elk} (‘every’, \textsc{neut}) & 38 \\
13. & \textit{blijkt} (‘turns out’) & 37 \\
14. & \textit{geen} (‘no’) & 35 \\
15. & \textit{je} (‘you’, \textsc{inform}) & 35 \\
16. & \textit{de} (‘the’, \textsc{masc/fem/pl}) & 32 \\
17. & \textit{van} (‘of’) & 32 \\
18. & \textit{worden} (‘become’) & 32 \\
19. & \textit{door} (‘by’) & 31 \\
20. & \textit{u }(‘you’, \textsc{form}) & 31 \\
21. & \textit{jij} (‘you’, \textsc{inform}) & 29 \\
22. & \textit{wordt} (‘becomes’) & 29 \\
23. & \textit{hoe} (‘how’) & 28 \\
24. & \textit{elke} (‘every’, \textsc{masc/fem}) & 27 \\
25. & \textit{massa's} (‘masses’) & 27 \\
26. & \textit{veel} (‘a lot of’) & 27 \\
27. & \textit{werkt} (‘works’) & 27 \\
28. & \textit{hier} (‘here’) & 26 \\
29. & \textit{waarom} (‘why’) & 25 \\
30. & \textit{ieder} (‘every’, \textsc{neut}) & 24 \\
31. & \textit{weinig} (‘a few’) & 23 \\
32. & \textit{aan} (‘to’, \textsc{prep}) & 22 \\
33. & \textit{iedere} (‘every’, \textsc{masc/fem}) & 21 \\
34. & \textit{voor} (‘for’) & 21 \\
35. & \textit{waar} (‘where’) & 21 \\ \hline
\end{tabular}
\caption{35 most frequent words for 1,000 generated sentences}
\label{tab:1000S_training}
\end{minipage}
\end{table*}

\begin{table}[h]
\centering
\begin{tabular}{lP{4cm}P{4cm}}
\hline
& \parbox[c][0.9cm][c]{4cm}{\centering \hspace{0.7cm} \textbf{NP}\newline \hspace{0.2cm} \textbf{training experiment}} & \parbox[c][0.9cm][c]{4cm}{\centering \hspace{0.7cm} \textbf{Sentence}\newline \hspace{0.2cm} \textbf{training experiment}} \\
\hline
\textbf{Number of word types} & 2,989 & 3,455\\
\textbf{Number of word tokens} & 30,101 & 57,732\\
\textbf{Number of hapax legomena} & 376  & 418\\
\textbf{Number of word types used twice} & 396 & 355\\
\hline
\end{tabular}
\caption{Lexical statistics of the generated 10,000 NP and sentences training data}
\label{tab:lex_stat_train_10000}
\end{table}

Accordingly, the follow-up experiment employed the larger data sets containing 10,000 NPs and sentences generated by Delilah. Table \ref{tab:lex_stat_train_10000} summarizes the lexical statistics for these data sets.  Similarly as for the previous experiment, the 35 most frequent words in both data sets are ranked in Table \ref{tab:10000NP_training} and Table \ref{tab:10000S_training} respectively: As shown, these 35 highest-ranked word types are generally function words. In the sample of 10,000 NPs, the first unequivocal content word, \textit{vergeten} (‘forgotten’), appears at rank 54/55 sharing its position with the function word \textit{bij} (‘at’, ‘near’). After rank 55, the amount of content words increases. Similarly, in the experiment with 10,000 sentence training exemplars generated by Delilah, the most frequent content words are \textit{vergeten} (‘forgotten’) at rank 11, \textit{werkt} (‘works’) at position 27, and \textit{morgens} (part of the archaic idiom \textit{’s morgens} ‘in the morning’) at rank 66. After rank 66, the number of content words increases. Notably, Delilah uses the word \textit{massa’s} (‘masses’) as a numeral, which is accurately detected by these acquisition procedures.

\begin{table*}[h!]
\begin{minipage}[l]{.4\linewidth}
\centering
\begin{tabular}[t]{@{}llll@{}}
\hline
\# & Word type & $f$ \\ \hline
1. & \textit{te} (‘at’, ‘to’, \textsc{comp}) & 1018 \\
2. & \textit{dat} (‘that’) & 658 \\
3. & \textit{niet} (‘not’) & 650 \\
4. & \textit{om} (‘to’) & 446 \\
5. & \textit{de} (‘the’, \textsc{masc/fem/pl}) & 291 \\
6. & \textit{er} (‘there’) & 271 \\
7. & \textit{geen} (‘no’) & 270 \\
8. & \textit{elk} (‘every’, \textsc{neut}) & 251 \\
9. & \textit{iedere} (‘every’, \textsc{masc/fem}) & 247 \\
10. & \textit{of} (‘or’, ‘whether’) & 244 \\
11. & \textit{elke} (‘every’, \textsc{masc/fem}) & 229 \\
12. & \textit{ieder} (‘every’, \textsc{neut}) & 223 \\
13. & \textit{alle} (‘all’) & 215 \\
14. & \textit{een} (‘a’, ‘one’) & 212 \\
15. & \textit{ik} (‘I’) & 194 \\
16. & \textit{massa's} (‘masses’) & 186 \\
17. & \textit{het} (‘the’, \textsc{neut:sg}) & 180 \\
18. & \textit{op} (‘on’) & 178 \\
19. & \textit{deze} (‘this’, \textsc{masc/fem}) & 172 \\
20. & \textit{weinig} (‘a few’) & 171 \\
21. & \textit{daar} (‘there’) & 163 \\
22. & \textit{aan} (‘to’, \textsc{prep}) & 161 \\
23. & \textit{veel} (‘a lot of’) & 155 \\
24. & \textit{hier} (‘here’) & 146 \\
25. & \textit{voor} (‘for’) & 144 \\
26. & \textit{dit} (‘this’, \textsc{neut}) & 140 \\
27. & \textit{zijn} (‘are’, ‘be’, ‘his’) & 130 \\
28. & \textit{door} (‘by’) & 122 \\
29. & \textit{in} (‘in’) & 120 \\
30. & \textit{menig} (‘many’, \textsc{sg}) & 118 \\
31. & \textit{menige} (‘many’, \textsc{pl}) & 111 \\
32. & \textit{je} (‘you’, \textsc{inform}) & 110 \\
33. & \textit{die} (‘that’, \textsc{masc/fem}) & 103 \\
34. & \textit{waar} (‘where’) & 97 \\
35. & \textit{sommige} (‘some’, \textsc{pl}) & 90 \\ \hline
\end{tabular}
\caption{35 most frequent words for the 10,000 generated NPs training data}
\label{tab:10000NP_training}
\end{minipage}\hfill\begin{minipage}[l]{.4\linewidth}\centering
\begin{tabular}[t]{@{}llll@{}}
\hline
\# & Word type & $f$ \\ \hline
1. & \textit{dat} (‘that’) & 1999 \\
2. & \textit{te} (‘at’, ‘to’, \textsc{comp}) & 1779 \\
3. & \textit{er} (‘there’) & 1370 \\
4. & \textit{niet} (‘not’) & 969 \\
5. & \textit{ik} (‘I’) & 729 \\
6. & \textit{hebben} (‘have’) & 599 \\
7. & \textit{blijkt} (‘turns out’) & 460 \\
8. & \textit{om} (‘to’) & 434 \\
9. & \textit{alle} (‘all’) & 409 \\
10. & \textit{zijn} (‘are’, ‘be’, ‘his’) & 386 \\
11. & \textit{vergeten} (‘forgotten’) & 385 \\
12. & \textit{elk} (‘every’, \textsc{neut}) & 330 \\
13. & \textit{de} (‘the’, \textsc{masc/fem/pl}) & 324 \\
14. & \textit{ieder} (‘every’, \textsc{neut}) & 317 \\
15. & \textit{iedere} (‘every’, \textsc{masc/fem}) & 317 \\
16. & \textit{door} (‘by’) & 315 \\
17. & \textit{geen} (‘no’) & 294 \\
18. & \textit{aan} (‘to’, \textsc{prep}) & 287 \\
19. & \textit{veel} (‘a lot of’) & 281 \\
20. & \textit{elke} (‘every’, \textsc{masc/fem}) & 280 \\
21. & \textit{hier} (‘here’) & 280 \\
22. & \textit{u} (‘you’, \textsc{form}) & 272 \\
23. & \textit{daar} (‘there’) & 269 \\
24. & \textit{op} (‘on’) & 268 \\
25. & \textit{weinig} (‘a few’) & 268 \\
26. & \textit{massa's} (‘masses’) & 262 \\
27. & \textit{werkt} (‘works’) & 255 \\
28. & \textit{deze} (‘this’, \textsc{masc/fem}) & 253 \\
29. & \textit{worden} (‘become’) & 240 \\
30. & \textit{jij} (‘you’, \textsc{inform}) & 238 \\
31. & \textit{je} (‘you’, \textsc{inform}) & 226 \\
32. & \textit{wordt} (‘becomes’) & 224 \\
33. & \textit{waar} (‘where’) & 216 \\
34. & \textit{een} (‘a’, ‘one’) & 213 \\
35. & \textit{is} (‘is’) & 210 \\ \hline
\end{tabular}
\caption{35 most frequent words for the 10,000 generated sentences training data}
\label{tab:10000S_training}
\end{minipage}
\end{table*}

\begin{figure*}[h!]
\centering
\includegraphics[width=1\textwidth]{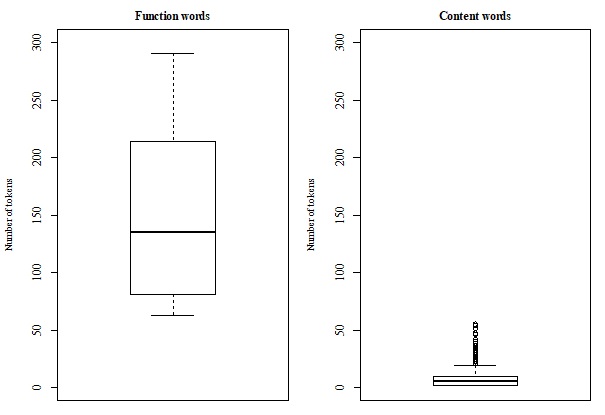}
\caption{Boxplots depicting function and content words for the 10,000 NPs training data}
\label{fig:10000NP_training}
\end{figure*}

\begin{figure*}[h]
\centering
\includegraphics[width=1\textwidth]{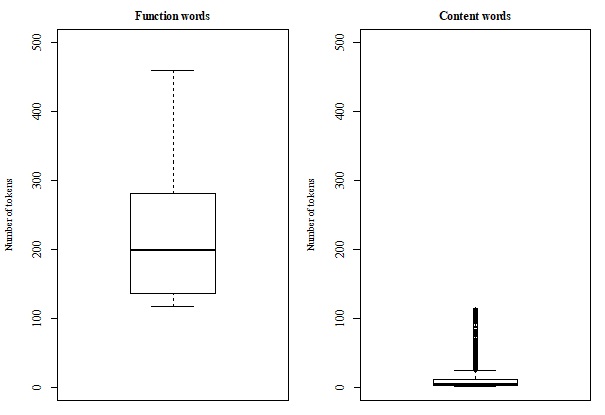}
\caption{Boxplots depicting function and content words for the 10,000 sentences training data}
\label{fig:10000S_training}
\end{figure*}

\section{Analyses}
\label{sec:Analyses}
For data sets of 10,000 exemplars, a more straightforward difference between function and content categories can be manually detected such that words with a frequency of more than 100 tokens tend to be function words, that is, more than 100/57,732 or 1.7 per mil for the sample of Delilah sentences. Therefore, based on these experimentations the MODOMA parameters determining the acquisition of functional and content categories are set in order that word types that have a frequency of more than 2 per mil, are specified as functional categories while all other word types are considered content words. In addition, this acquisition procedure is executed only once after inputting at least 10,000 exemplars. These parameters are set conservatively, that is, in such a way that the chances of falsely classifying a word as functional category are smaller than assigning a specification as content word: As content word is the default category, each word should be considered a content word unless there is strong evidence for the opposite conclusion. The parameter values resulting from this initial experiment have been configured as the default settings for the MODOMA. However, these parameters can be adjusted depending on the experiment and taking into account previous experiments and linguistic theories. 

Using the results of acquisition based on these settings, the boxplots in Figure \ref{fig:10000NP_training} and Figure \ref{fig:10000S_training} have been made taking into account the frequency of both word types and these groupings suggest that for the NP as well as the sentence experiment the function words are distributed around other central values than the content words.\footnote{The boxplots for the function words displayed outliers attested with a very higher frequency. Although these further confirm the patterns found that function words correspond to high rankings, these extreme values have been omitted as they would render the figures unreadable.} This provides further support for the research hypothesis and the selected default parameter settings. Crucially, although the data are continuous, the acquired linguistic knowledge is discrete, that is, a word is either classified as a content or a functional category. These abstract categories explicitly specify the corresponding lexical items using a binary rule-based grammar formalism. This allows the daughter agent to apply the acquired feature-value pairs to generate and parse utterances based on unification during the conversation with the mother agent (cf. \citename{Pollard:1986} 1986, \citename{Pollard:1994} 1994 as well as Delilah for examples of unification-based language models, which do not implement language acquisition).

To quantitatively assess whether the acquired knowledge of the daughter agent matches the grammatical knowledge of the mother agent, a comparison has been made between the explicit grammatical specifications of both language models. Delilah does not directly specify for functional and content categories. Nonetheless, the mother language model employs categories such as transitive verb, count noun, or coordinating conjunction to process language, which correspond to function and content words. Therefore, the classifications in the grammar of the daughter resulting from the language acquisition simulations can be compared to the specifications in Delilah’s core lexicon. This database contains the basic grammatical knowledge employed by the mother agent to generate the input exemplars to the language acquisition procedures. Fisher’s exact tests have been executed to examine the relationship between a classification as function and content category by the mother and daughter language models. For all training experiments the results indicated there was a significant association between these variables, $p < 0.001$ (two-tailed). Further analyses revealed that in addition to matching classifications of both language models the mismatching classifications were predominantly function words according to the mother language model classified as content categories by the daughter agent.

\section{Testing the suggested default settings using other data sets}
\label{sec:Testing}
To validate the parameter settings determined during the training sessions, the MODOMA has been requested to acquire function and content categories based on data sets not used to configure the parameter settings. To this end, two new data sets with respectively 10,000 NPs and sentences generated by Delilah have been created. Similarly as with respect to the training sets, the acquisition procedures for the test sets resulted in lists containing all word types in the data sets ranked in the order of the most frequent to the least frequent words as intermediate analyses. In accordance with the results of experimentation on the training data, the most frequent words correspond mostly to function words while the remainder of the words tend to be representative of content categories (according to more conventional grammar models). The 35 most frequent types occurring in these new data sets have been listed in Table \ref{tab:10000NP_test} for the 10,000 NP sample and Table \ref{tab:10000S_test} for the exemplars representing 10,000 sentences.

\begin{table*}
\begin{minipage}[l]{.4\linewidth}
\centering
\begin{tabular}[t]{@{}llll@{}}
\hline
\# & Word type & $f$ \\ \hline
1. & \textit{te} (‘at’, ‘to’, \textsc{comp}) & 1025 \\
2. & \textit{niet} (‘not’) & 633 \\
3. & \textit{dat} (‘that’) & 631 \\
4. & \textit{om} (‘to’) & 451 \\
5. & \textit{geen} (‘no’) & 304 \\
6. & \textit{de} (‘the’, \textsc{masc/fem/pl}) & 292 \\
7. & \textit{er} (‘there’) & 261 \\
8. & \textit{ieder} (‘every’, \textsc{neut}) & 253 \\
9. & \textit{elk} (‘every’, \textsc{neut}) & 237 \\
10. & \textit{iedere} (‘every’, \textsc{masc/fem}) & 231 \\
11. & \textit{elke} (‘every’, \textsc{masc/fem}) & 224 \\
12. & \textit{of} (‘or’, ‘whether’) & 215 \\
13. & \textit{een} (‘a’, ‘one’) & 203 \\
14. & \textit{alle} (‘all’) & 200 \\
15. & \textit{massa's} (‘masses’) & 194 \\
16. & \textit{ik} (‘I’) & 193 \\
17. & \textit{op} (‘on’) & 178 \\
18. & \textit{aan} (‘to’, \textsc{prep}) & 173 \\
19. & \textit{deze} (‘this’, \textsc{masc/fem}) & 172 \\
20. & \textit{het} (‘the’, \textsc{neut:sg}) & 167 \\
21. & \textit{veel} (‘a lot of’) & 162 \\
22. & \textit{weinig} (‘a few’) & 160 \\
23. & \textit{voor} (‘for’) & 145 \\
24. & \textit{hier} (‘here’) & 143 \\
25. & \textit{die} (‘that’, \textsc{masc/fem}) & 135 \\
26. & \textit{menige} (‘many’, \textsc{pl}) & 131 \\
27. & \textit{door} (‘by’) & 130 \\
28. & \textit{zijn} (‘are’, ‘be’, ‘his’) & 130 \\
29. & \textit{dit} (‘this’, \textsc{neut}) & 121 \\
30. & \textit{in} (‘in’) & 120 \\
31. & \textit{daar} (‘there’) & 114 \\
32. & \textit{menig} (‘many’, \textsc{sg}) & 114 \\
33. & \textit{je} (‘you’, \textsc{inform}) & 109 \\
34. & \textit{drie} (‘three’) & 102 \\
35. & \textit{twee} (‘two’) & 89 \\ \hline
\end{tabular}
\caption{35 most frequent words for the 10,000 generated NPs test data}
\label{tab:10000NP_test}
\end{minipage}\hfill\begin{minipage}[l]{.4\linewidth}\centering
\begin{tabular}[t]{@{}llll@{}}
\hline
\# & Word type & $f$ \\ \hline
1. & \textit{dat} (‘that’) & 1937 \\
2. & \textit{te} (‘at’, ‘to’, \textsc{comp}) & 1829 \\
3. & \textit{er} (‘there’) & 1353 \\
4. & \textit{niet} (‘not’) & 929 \\
5. & \textit{ik} (‘I’) & 714 \\
6. & \textit{hebben} (‘have’) & 570 \\
7. & \textit{om} (‘to’) & 448 \\
8. & \textit{blijkt} (‘turns out’) & 422 \\
9. & \textit{zijn} (‘are’, ‘be’, ‘his’) & 396 \\
10. & \textit{vergeten} (‘forgotten’) & 389 \\
11. & \textit{alle} (‘all’) & 354 \\
12. & \textit{de} (‘the’, \textsc{masc/fem/pl}) & 330 \\
13. & \textit{geen} (‘no’) & 330 \\
14. & \textit{ieder} (‘every’, \textsc{neut}) & 315 \\
15. & \textit{door} (‘by’) & 304 \\
16. & \textit{elke} (‘every’, \textsc{masc/fem}) & 301 \\
17. & \textit{u} (‘you’, \textsc{form}) & 294 \\
18. & \textit{elk} (‘every’, \textsc{neut}) & 293 \\
19. & \textit{aan} (‘to’, \textsc{prep}) & 278 \\
20. & \textit{hier} (‘here’) & 276 \\
21. & \textit{jij} (‘you’, \textsc{inform}) & 272 \\
22. & \textit{iedere} (‘every’, \textsc{masc/fem}) & 268 \\
23. & \textit{daar} (‘there’) & 267 \\
24. & \textit{op} (‘on’) & 259 \\
25. & \textit{veel} (‘a lot of’) & 258 \\
26. & \textit{weinig} (‘a few’) & 250 \\
27. & \textit{je} (‘you’, \textsc{inform}) & 249 \\
28. & \textit{massa's} (‘masses’) & 249 \\
29. & \textit{deze} (‘this’, \textsc{masc/fem}) & 238 \\
30. & \textit{een} (‘a’, ‘one’) & 236 \\
31. & \textit{worden} (‘become’) & 233 \\
32. & \textit{werkt} (‘works’) & 221 \\
33. & \textit{wordt} (‘becomes’) & 209 \\
34. & \textit{waar} (‘where’) & 207 \\
35. & \textit{is} (‘is’) & 206 \\ \hline
\end{tabular}
\caption{35 most frequent words for the 10,000 generated sentences test data}
\label{tab:10000S_test}
\end{minipage}
\end{table*}

\begin{table}
\centering
\begin{tabular}{lcc}
\hline
 & \textbf{NP test experiment} & \textbf{Sentence test experiment}\\
\hline
\textbf{Number of word types} & 3,006 & 3,442\\
\textbf{Number of word tokens} & 30,129 & 57,118\\
\textbf{Number of hapax legomena} & 379  & 411\\
\textbf{Number of word types used twice} & 380 & 372\\
\hline
\end{tabular}
\caption{Lexical statistics of the generated 10,000 NP and sentences test data}
\label{tab:lex_stat_test_10000}
\end{table}

\begin{figure*}
\centering
\includegraphics[width=1\textwidth]{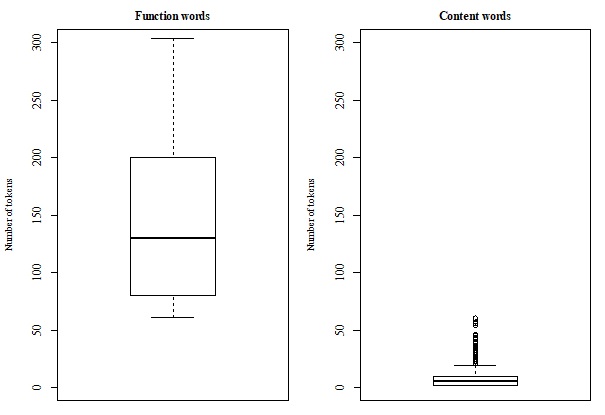}
\caption{Boxplots depicting function and content words for the 10,000 NPs test data}
\label{fig:10000NP_test}
\end{figure*}

\begin{figure*}[h!]
\centering
\includegraphics[width=1\textwidth]{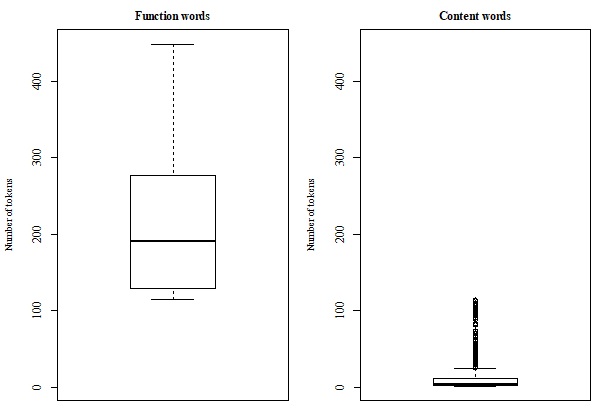}
\caption{Boxplots depicting function and content words for the 10,000 sentences test data}
\label{fig:10000S_test}
\end{figure*}

Table \ref{tab:lex_stat_test_10000} summarizes the distribution of word types, word tokens, and hapax legomena for the test sets containing 10,000 NPs and sentences. For the test set containing 10,000 NPs, the first content word in the corresponding list is \textit{vragen} (‘questions’, \textsc{n}, ‘question’, \textsc{v}) at rank 56, which is attested with a frequency of 56. Therefore, the suggested default setting determined based on the experimentations with the training data, which predicts items occurring with a frequency greater than approximately 60 to be function words, is confirmed by the experiment assessing the NP test set. Similarly, with respect to the list constructed based on the 10,000 sentences test set, the first unequivocal content word is \textit{weer} (‘again’, ‘weather’) at rank 50 with a frequency of 130 while the parameter settings configured by the training experiment predict the threshold to be at a frequency of 114 tokens. Thus, both test sets provide further support for the suggested parameter settings resulting from the previous experimentations. 

Concomitantly, the boxplots corresponding to these data sets visualize the categories resulting from these parameter settings in respectively Figure \ref{fig:10000NP_test} and Figure \ref{fig:10000S_test}. Similarly to the results of the training data, it can be observed that for each experiment  the groups of function words and content words cluster around different central values. Taking into account these classifications of words, the corresponding lexical entries in the MODOMA daughter grammar are subcategorized as either functional or content categories. Moreover, for the results of the test sets the classifications acquired by the daughter agent have been compared to the grammatical categories employed by the mother language model and Fisher’s exact tests have been executed to assess this relationship. Similar to the training experiments, a significant association was found between the categories employed by the mother agent and the categories acquired by the daughter agent, $p < 0.001$ (two-tailed). Thus, the MODOMA daughter agent has learned non-trivial grammatical classifications represented by the currently acquired language model, which it can employ productively to generate novel utterances and analyze data generated by the mother agent.

\section{Conclusions and suggestions for future research}
\label{sec:Conclusions_suggestions}
This paper discussed how functional and content categories can be acquired by a computational multi-agent language acquisition laboratory involving an adult language model, Delilah, and a daughter agent. The daughter language model has been constructed for the purpose of the MODOMA and employs explicit graph representations of the acquired grammatical knowledge to generate and parse utterances, which become more specific as a result of language acquisition. Conversely, Delilah consists of a parser and generator that have been independently developed by \citeasnoun{Cremers:1995} to implement a grammar model of Dutch, and employs this model to output and parse utterances \citeaffixed[for an elaborate discussion of the attributes underlying this language model]{Cremers:2014}{cf.}. Thus, the MODOMA framework requires taking a new approach to research with respect to Delilah: in addition to constructing and improving Delilah to produce utterances and provide parses, the properties of the output of Delilah should be studied as well to enable the daughter to detect the patterns needed for acquisition. To this end, analyses and techniques from for instance the fields of linguistics, psycholinguistics, and corpus linguistics could be applied to gain further insights in the machine-generated output of this language model similarly to how they have been applied to human-generated language data. Taking into account the current rise of applications of large language models, evaluating whether machine-generated language exhibits similar patterns to language produced by humans has become increasingly relevant. This paper provides the results of experiments exploring this approach. 

Interestingly, the collection of Delilah utterances generated during the conversation with the daughter agent contains independent exemplars while corpora typically form a cohesive collection of subsequent sentences. Nonetheless, similar patterns corresponding to grammatical phenomena can be detected with respect to both types of linguistic data. This might indicate that these patterns are rather a characteristic of the individual utterances contained in these collections than of the corpora. Crucially, the study presented by this paper reveals that the patterns regarding the divergent frequency distributions of content and functional categories, which are well-established for human language data in the corpus linguistics literature \citeaffixed{Powers:1998}{cf.} are also found in Delilah’s output. As Delilah is not configured to reproduce these statistical patterns in its output utterances, the found distribution is a consequence of the specified grammar model.

The initial experiments have been used to set the parameters of the MODOMA acquisition system related to the acquisition of function and content categories, that is, (1) the amount of input data that is required before executing the functional and content categories acquisition procedures, and (2) the threshold for determining whether a word should be classified as either a functional or content category. It was assessed how many input exemplars need to be processed to detect this statistical tendency by conducting experiments on 1,000 and 10,000 NP and sentence training sets. This study revealed that only in data sets containing 10,000 exemplars generated by Delilah a frequency distribution differentiating function and content words can be sufficiently detected. In addition, the threshold value was set to 2 per mil to convert the differences between the distributions of function and content words to discrete linguistic knowledge.

Moreover, these parameter settings have been applied to two test sets containing respectively 10,000 NPs and 10,000 sentences generated by Delilah as well. These experiments confirmed the parameter settings whose values were determined based on the training sets, by similarly allowing for the acquisition of content and functional categories. Figure \ref{fig:10000NP_test} and Figure \ref{fig:10000S_test} visualize the clusters resulting from the acquisition procedures performed on these test sets. Although the input data are noisy and form a continuous distribution, compare Figure \ref{fig:10000NP_training} and Figure \ref{fig:10000S_training}, the unsupervised acquisition procedures result in a discrete knowledge-based grammar: each lexical item is classified as either a functional or content category and these properties are specified by feature-value pairs subcategorizing the corresponding lexical graphs in the grammar. Thus, these newly acquired grammatical categories can be used productively to generate and parse utterances by the daughter agent as they specify the combinatorial properties of lexical items during unification.

The combination of properties of the MODOMA system contributes to the field by providing a laboratory approach to modeling language acquisition. Both the adult and child agents interact as integral components of the system. Moreover, the system is fully parameterized, and the acquired grammatical knowledge is explicitly represented and can be consulted. Crucially, as this grammatical knowledge is explicitly represented by the grammar model, the linguistic knowledge and processing of the language model can be straightforwardly consulted by researchers. This design expands the possibilities for simulating language acquisition as many language models that implement language learning, do not employ explicit knowledge representations. Accordingly, these representations were used to assess whether the grammatical categories employed by the mother language model to generate the input to language acquisition are associated with those acquired by the daughter agent, which is significant for the results of all training and test experiments ($p<0.001$).  This study carried out using the MODOMA supports the research hypothesis that taking into account the differences between humans and artificial intelligence systems the MODOMA framework enables the unsupervised acquisition of discrete grammatical categories such as function and content words by employing statistical techniques. Thus, this result substantiates that a MODOMA can provide a computational laboratory model for simulating language acquisition experiments to study linguistic phenomena.

An important contribution of the study presented in this paper is the validation of the feasibility of a multi-agent computational laboratory for first language acquisition. The application of the principle of Zipf’s law thus serves as an essential tool for evaluating the MODOMA approach to modeling language acquisition. Crucially, the experiments designed to acquire function and content categories, using a hybrid approach involving statistical analyses that result in discrete categories, incorporate all the elements found in more complex experiments. As a result, the findings of this study lay a solid foundation for future investigations. For instance, this research paves the way for further studies that can expand and refine the current experiments: future research extending the function and content word experiments could explore the effects of new parameters on the grammar acquired by the daughter agent, the utterances produced by the daughter agent, and the interaction with the mother agent. Examples of such parameters include the number of times the functional and content category acquisition process is executed as well as whether previously acquired categories should be changed when additional data is processed. Another relevant avenue for further study is to investigate the impact of feedback from the mother agent to the daughter agent and its effect on language acquisition, the generated utterances and the interaction between the adult and child agent. Furthermore, the success of the current experiments provides the opportunity to test more complex acquisition procedures. Extending this framework, the acquisition of additional and more intricate grammatical phenomena should be modeled using the MODOMA approach. For example, \citeasnoun{Shakouri:2025} demonstrate that grammatical categories such as nouns, verbs, and prepositions can be acquired under similar experimental conditions as the function and content words experiment. Building on the results of this study, Shakouri et al. (2025) employ more advanced statistical methods, such as hierarchical agglomerative clustering, to acquire discrete grammatical categories in an unsupervised manner. Thus, the results of this study provide a foundational contribution for the continued exploration and development of this line of research.

\section*{Acknowledgements}
\label{Acknowledgements}
The authors would like to thank Maarten Hijzelendoorn for his assistance. This paper presents results of the MODOMA project, which was funded by the Dutch Research Council (NWO).

\bibliographystyle{clin} 
\bibliography{Shakouri_et_al_knowledge_based_lang_model}
\appendix

\newpage
\section{Glossary of abbreviations}
\label{appendix:A}
\setcounter{table}{0}
\renewcommand{\thetable}{A.\arabic{table}}
\begin{table}[!h]
\begin{center}
\begin{tabular}{ l l }
\hline
\textbf{Abbreviation} & \textbf{Description} \\ \hline
\textsc{comp} & complementizer \\
\textsc{fem} & feminine \\
\textsc{form} & formal \\
\textsc{inform} & informal \\
\textsc{masc} & masculine \\
\textsc{n} & noun \\
\textsc{neut} & neuter \\
\textsc{pl} & plural \\
\textsc{prep} & preposition \\
\textsc{sg} & singular \\
\textsc{v} & verb \\ \hline
\end{tabular}
\caption{Glossary of abbreviations for grammatical terminology}
\label{tab:app_gloss}
\end{center}
\end{table}

\end{document}